\documentclass[conference]{IEEEtran}
\IEEEoverridecommandlockouts
\usepackage{cite}
\usepackage{amsmath,amssymb,amsfonts}
\usepackage{graphicx}
\usepackage{textcomp}
\usepackage{xcolor}
\usepackage{soul} % highlighting 
\usepackage{enumitem} % Margins in enumitem

\usepackage{multirow}
\usepackage{tabularx}
\usepackage{bm}
\usepackage{subfig} % side by side images
\usepackage{amsmath}
\usepackage{amsfonts}
\usepackage{booktabs}
\usepackage{balance}
\usepackage[english]{babel}
\usepackage[utf8]{inputenc}
\usepackage[noend]{algpseudocode}
\usepackage{centernot}
\usepackage{adjustbox}
\usepackage{multirow}
\usepackage{todonotes}

\usepackage{xspace} % Needed for et al.
\newcommand{\ie}{\emph{i.e.,}\xspace}
\newcommand{\eg}{\emph{e.g.,}\xspace}

\newcommand{\etal}{\emph{et~al.}\xspace} 

\def\BibTeX{{\rm B\kern-.05em{\sc i\kern-.025em b}\kern-.08em
    T\kern-.1667em\lower.7ex\hbox{E}\kern-.125emX}}

\begin{document}

\title{Uncertainty-Aware Multiple Instance Learning from Large-Scale Long Time Series Data}

\author{Anonymous}

\author{\IEEEauthorblockN{Yuansheng Zhu, Weishi Shi, Deep Shankar Pandey, Yang Liu, Xiaofan Que, Daniel E. Krutz, and Qi Yu}
\IEEEauthorblockA{Rochester Institute of Technology\\
Rochester, NY, USA \\
\{yz7008, ws7586, dp7972, yl4070, xq5054, dxkvse, qyuvks\}@rit.edu}
}

\maketitle

\begin{abstract}

We propose a novel framework to classify large-scale time series data with long duration. Long time series classification (L-TSC) is a challenging problem because the data often contains a large amount of irrelevant information to the classification target. The irrelevant period degrades the classification performance while the relevance is unknown to the system. This paper proposes an uncertainty-aware multiple instance learning (MIL) framework to identify the most relevant period automatically. 
The predictive uncertainty enables designing an attention mechanism that forces the MIL model to learn from the possibly discriminant period. Moreover, the predicted uncertainty yields a principled estimator to identify whether a prediction is trustworthy or not. We further incorporate another modality to accommodate unreliable predictions by training a separate model based on its availability and conduct uncertainty aware fusion to produce the final prediction. Systematic evaluation is conducted on the Automatic Identification System (AIS) data, which is collected to identify and track real-world vessels. Empirical results demonstrate that the proposed method can effectively detect the types of vessels based on the trajectory and the uncertainty-aware fusion with other available data modality (Synthetic-Aperture Radar or SAR imagery is used in our experiments) can further improve the detection accuracy.

\end{abstract}


%-------------------------------------------------------------------------------------------------------------------
\section{Introduction}
\label{sec: introduction}

Time series classification (TSC) is a central component for a myriad of classification tasks, ranging from health care\cite{rajkomar2018scalable}, power system~\cite{susto2018time} to trajectory classification~\cite{xing2019personalized}. Deep Neural Network (DNN) is a popular choice for TSC problems due to its I) robust performance and II) appropriateness for big data~\cite{fawaz2019deep}. However, longer time series poses novel challenges as it may contain many irrelevant parts (\eg, in a fishing vessel trajectory, only ``engaged in fishing'' parts are relevant and unique to this specific ship type). As a result, using the entire long time series to train a DNNs model may eventually lead to overfitting the noises (\ie parts of the trajectory irrelevant to the classification target), degrading the model's generalization performance~\cite{goodfellow2016deep}. Additionally, predictions made without identifying the discriminant sub-sequence are less valuable because domain knowledge cannot properly justify the results. Unfortunately, the irrelevant parts are universal and redundant in long time series data and hard to eliminate directly (\ie, relevance are unknown to the system). Lack of interpretability hinders the application of TSC methods in broader domains.

The problem of data noise in TSC is traditionally addressed by
``split and search" order. These methods first split the time series into multiple sub-sequences and then exhaustively search the most discriminant sub-sequences. These discriminant sub-sequences are commonly referred to as {\em shapelets}, and measurement of shapelets can be based on information gain~\cite{ye2009time, mueen2011logical},  F-Stat~\cite{hills2014classification}, or Kruskall-Wallis~\cite{lines2012alternative}. Although some works (\eg~\cite{wistuba2015ultra}) have been proposed to reduce the computational cost of the exhaustive search process, these shapelets-based methods become much less effective with the increase of the volume of the time series data.

Our work is novel in that we uniquely train the DNNs models while adapting the idea of ``identifying shapelets while classifying the sequence". More specifically, we formalize shapelets identification as a multiple instance learning (MIL) problem~\cite{maron1998framework}, where the whole sequence is referred to as a bag, while each instance is a sub-sequence. The bag label is the entire sequence's label, while the instance label indicates whether a sub-sequence is shapelets or not. The MIL's bag assumption fits squarely into our problem, and its instance level prediction further adds interpretability to explain why certain bag-level prediction is made. More importantly, as powerful weakly supervised learning paradigm, MIL avoids significant pre-processing work, which is expensive and tedious for big data. Instance-level predictions can be conveniently aggregated into the bag-level prediction, which largely simplifies the training procedure compared to traditional shapelets based methods. Our novel contribution lies in how to identify the shapelets to make the instance-level classifier robust to the \textit{label noise}.

This work also provides a novel uncertainty-guided loss function, adjusting the weight of each instance in a bag to make the DNNs focus upon the instance that is more likely to align with the bag label (\eg being shapelets). We use the Bayesian Neural Network (BNN) to obtain the predictive uncertainty and then use uncertainty to design the attention mechanism. This uncertainty knowledge benefits the training process and the inference phase. We further augment the MIL model with {\em uncertainty-aware multimodal data fusion} to effectively address data noise.  We also introduce a second modality, image data, to improve our overall prediction confidence. Two models are individually trained, and the prediction confidence is used as the weight during prediction fusion, adaptively adjusting the weights of two models' prediction. 

We demonstrate the ability of our work through several real-world aquatic vessel classification tasks, using trajectory information from the \emph{Automatic Identification System} (AIS) vessel tracking system. This is accomplished through a comparative analysis against existing state-of-the-art techniques. We found our process to be more effective than previous state-of-the-art methods.
We summarize our contributions as follows:

\begin{enumerate}[leftmargin=.5cm]
    \item \textbf{Overall framework: }We develop a novel general framework for large-scale sequential data classification from the MIL perspective. Our uncertainty-aware MIL based framework is interpretable, accurate, and computationally efficient.

    \item \textbf{Uncertainty quantification:} We perform uncertainty quantification to yield a more accurate estimator of shapelets while providing a more principled way for multimodal prediction fusion. The predictive uncertainty can also benefit the decision-making process to make the human experts aware of the machine's confidence in its prediction. In that case, human experts can take further actions to verify the low confident prediction.

    \item \textbf{Real-world applications: }We validate our framework through real-world ship type classification problems using large-scale AIS trajectory data. While one modality may be sub-optimal, we leverage the SAR imagery to boost the classification accuracy. This application can be used to identify Illegal, Unregulated \& Unreported (IUU) Fishing behaviour, contributing to protecting the ocean environment, maintaining resource sustainability, and preventing economic loss.
\end{enumerate}

The rest of the paper is organized as follows. Section~\ref{sec: ProblemDefinition} introduces the definition of TSC task and how we cast it to a MIL problem. Section~\ref{sec:proposedTechnique} describes the proposed method in details. Section~\ref{sec: relatedworks} gives an overview of related works. Section~\ref{sec: Evaluation and Results} presents the evaluation setting and results, and Section~\ref{sec: conclusion} concludes the paper.

\begin{figure}
\centering

\subfloat[SAR Image of a fishing vessel. \label{fig: ExampleSarImage}]{\includegraphics[width=0.45\linewidth,height=2.8cm]{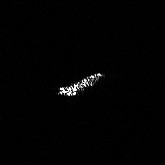}} \quad
\subfloat[Trajectory of a Fishing vessel based on AIS data (visualized with QGIS).\label{fig: ExampleAISTrajectory}]{\includegraphics[width=0.45\linewidth,height=2.8cm]{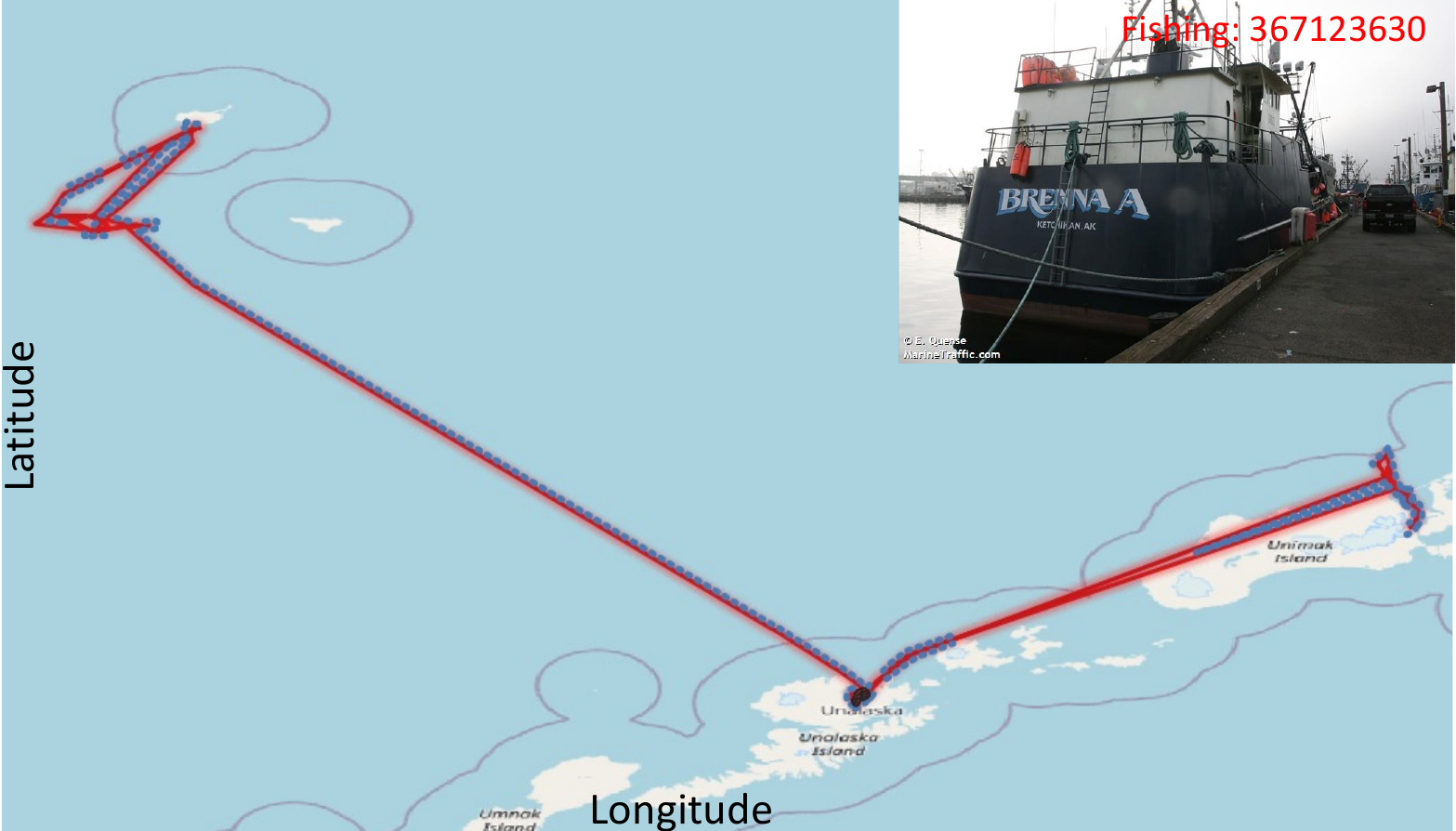}} \quad \\
\caption{Example SAR and AIS data modalities.}
\label{fig: ExampleSARandAISImages}
\vspace{-2mm}
\end{figure}

\section{Time Series Data Classification via MIL} 
\label{sec: ProblemDefinition}

This section presents the proposed MIL model for general long time series data classification with our use case with the AIS trajectory data.

%---------------------------------------------------------------
\subsection{MIL for Time series Data Classification}
\noindent\textbf{Time series dataset.} We use $\mathcal{D}=\{X,Y\}$ to denote a dataset containing $|\mathcal{D}|$ multivariate time-series samples. Each sample data $X$ contains $M$ time variables, and each variable contains $T$ values that are arranged in the temporal order. Formally, $X$ is a $M$ by $T$ matrix: $X\in\mathbb{R}^{M\times T}$, where each row denotes a time variable over all $T$ timestamps, and each column denotes all time variables at one timestamp. For example, $4$ variables [Latitude, Longitude, Speed Over Ground (SOG), Course Over Ground (COG)], can represent the information of a point in a vessel's trajectory. Correspondingly, the trajectory of one vessel over $100$ timestamps can be represented by a matrix of size $4\times 100$.
The series length $T$ may vary from sample to sample in real-world cases. Our method can handle varied series lengths, but we use a uniform $T$ to avoid notation clutter.

\vspace{2mm}\noindent\textbf{Multiple instance learning.} Unlike typical classification scenarios, where training samples are independent, training samples under MIL formulation exhibit temporal and/or spatial dependencies and subsequently are arranged into bags. For TSC problems, we treat each full sequence $X$ as a bag and use a sliding window to split a bag into multiple instances (sub-sequences), denoted by $\bm{x}$. The bag label $Y$ is the entire time sequence label, and the instance label $y$ indicates whether an instance corresponds to a shapelet or not. The relationship between bag label and instance label can be formatted as
\begin{align}
\label{eq:1}
    Y = 
    \left\{
   \begin{array}{ll} 
     0, & \text{if } \sum_{n}y_{n} = 0, \\
     1, & \text{otherwise}.
      \end{array}
   \right.
\end{align}
Eq.~\eqref{eq:1} implies that a positive bag ($\{X; Y=1\}$) contains at least one positive instance ($\{\bm{x}; \exists n \in [1, N], \text{s.t. } y_{n}=1\}$), and all instances are negative ($y_{n}=0, n=0,\ldots,N$) if the corresponding bag is negative ($Y=0$).

We propose to first learn a classifier that projects the instance feature to an instance prediction $f$: $\bm{x} \longrightarrow [0,1]$. Afterwards, we aggregate the instance-level prediction for bag-level prediction, as $\hat{Y} = \text{Aggregate}\{\hat{y}_{n}\}_{n=1}^{N}$. We assign each instance a pseudo instance label $\tilde{y}$ for each instances, which is inherited from their corresponding bag label, such that we can build a objective function to optimize the instance level classifier. Note that this MIL assumption reflects that only a portion of sub-sequences is discriminate, which are the positive instances. As indicated by eq.~\eqref{eq:1}, this pseudo instance label $\tilde{y}$ is not consistent with the true instance label when positive, \eg $\tilde{y} = 1$ can be either a true positive ($y=1|\tilde{y}=1$) or a false positive ($y=0|\tilde{y}=1$). To address this noisy label issue, we proposed uncertainty-aware MIL framework.

\section{Uncertainty-Aware Shapelets Identification and Data Fusion}
\label{sec:proposedTechnique}

\begin{figure*}[t]
\centering
\includegraphics[width=0.95\linewidth]{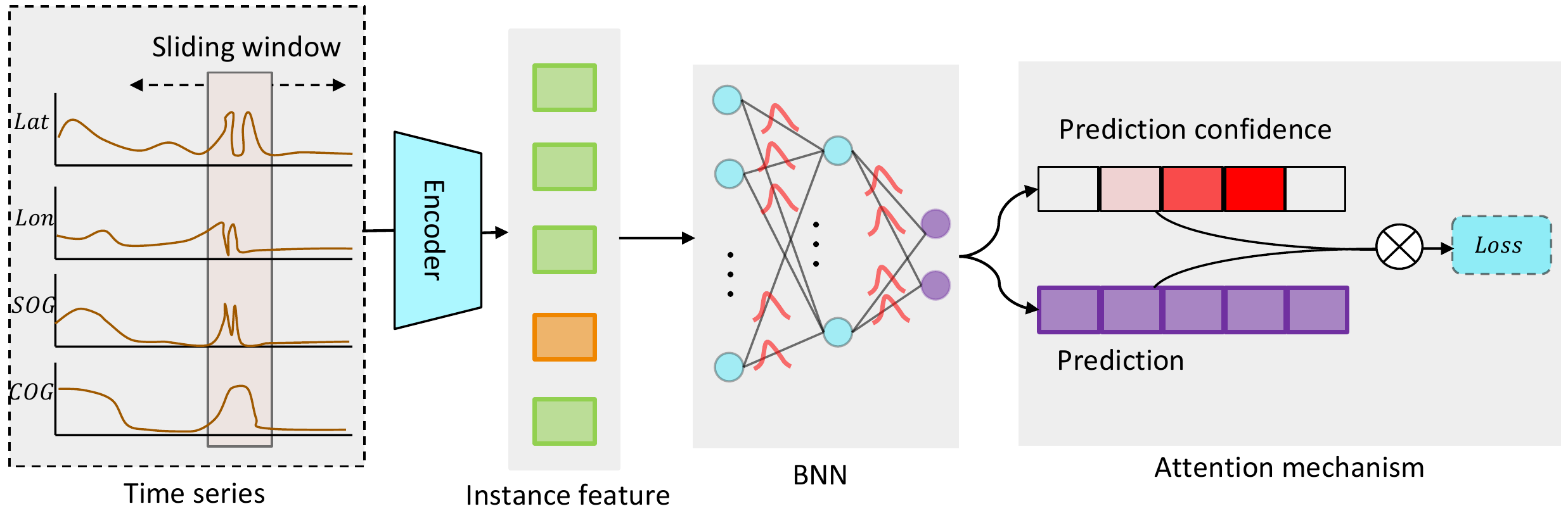} \quad
\caption{Overview of the framework. The feature encoder takes the time series data and outputs the feature for each instance. Then, the BNN outputs the mean prediction and variance. The variance is used for computing the predictive confidence, which yields an indicator for possibly positive instance. }
\label{fig:framework}
\vspace{-2mm}
\end{figure*}

Our key idea is to train an AIS model with instances whose pseudo labels $\tilde{y}$ are most likely to be true. Section~\ref{Overview of framework} gives an overview of our framework, and Section~\ref{sec: Uncertaintyquantification} and Section~\ref{subsec:loss} present two key building blocks of the framework. When the DNN makes uncertain predictions (\eg the AIS data quality is low for some samples), we leverage another modality (SAR). Section~\ref{Uncertainty-aware fusion.} demonstrates how we adaptively fuse the two modalities and improve the classification performance.

\subsection{Framework Overview}
\label{Overview of framework}
Figure~\ref{fig:framework} shows the overview of our AIS model. The framework is composed of a feature encoder and a BNN classification head.  We first use a sliding window to segment the entire time series into multiple non-overlap instances during every iteration. Then, we use an encoder to extract the temporal information of individual channels and the inter-channel relationships for each instance. We use the architecture from Tapnet~\cite{zhang2020tapnet} to build the encoder, which is composed of a long short-term memory network (LSTM) and d a multi-layer convolutional network (CNN). The encoder learns a low-dimension feature for each instance. Next, the instance features are fed to a BNN head, which outputs predictions and predictive confidence for every instance. Finally, through an attention mechanism, we compute the attention for each instance and acquire the loss. The encoder and BNN are trained in an end-to-end fashion. 

\subsection{Uncertainty Quantification}
\label{sec: Uncertaintyquantification}

In this work, we employ BNNs~\cite{blundell2015weight} as they can be easily extended to our classifier network, only double the number of parameters to learn and provide accurate uncertainty estimates without sacrificing the predictive performance. More specifically, instead of treating the parameters of the final classification head ($w)$ as the point estimates, we consider a distribution $p(w|\mathcal{D})$ for the classification head parameters. We then leverage variational inference to learn an optimal variational distribution of the weight parameters that best approximates the true posterior. To this end, we want to train the model such that the variational posterior distribution $q_\theta (w|\mathcal{D})$ is as close as possible to the true posterior $p(w|\mathcal{D})$. This objective is equivalent to solving the following optimization problem:
\begin{align}
    \theta^{*} &= \arg \min_\theta KL[q_\theta(w|\mathcal{D})||p(w|\mathcal{D})] \\
    &= \arg \min_\theta KL[q_\theta(w|\mathcal{D})||p(w)] - E_{q(w|\theta)}[\log p(\mathcal{D}|w)]
\end{align}
where $KL(\cdot||\cdot)$ is the KL-divergence between the two distributions. The optimization problem is analytically intractable. To address the intractability, we approximate the loss function using Monte-Carlo sampling as 
\begin{align}
\label{eq:7}
    F(\mathcal{D},\theta) = \sum_{j=1}^J \log q(w^{i}|\theta) - \log P(w^{i}) - \log P(\mathcal{D}|w^{i})
\end{align}
Here, we draw $J$ Monte-Carlo samples $w^{j} \sim q_{\theta}(w|\mathcal{D})$ from the variational posterior to approximate the loss. We optimize this approximated loss and train the model to learn the classifier head parameters. During the test phase, we take multiple weight samples (say $J$ samples) and make a prediction for the same input using each weight sample. We take the average of the $J$ predictions to make a final prediction, and a large variance of the predictions indicates a high uncertainty prediction.

Formally, for a $i^{th}$ instance, $\bm{x}^{i}$, we have $J$ predictions by running multiple forward passes. The assembled prediction $\bar{y}^{i}$ and predictive confidence $\hat{c}$ are given by
\begin{align}
    \bar{y}^{i} = \frac{1}{J}\sum_{j=1}^{J} \hat{y}^{i}, \hat{c}^{i} = \left(1 -  \text{Var}(\{\hat{y}^{i}_{j}\}_{j=1}^{J})\right)^{k},
\end{align}
where $k$ is the hyper-parameter controlling the gap between the confident instances and less confident ones. 

After acquiring the instance level prediction and confidence, we can attain the bag-level prediction by aggregate the bag level prediction and confidence by taking the mean of top instances, \ie instances with largest $\bar{y}$.

\subsection{Uncertainty-Driven Objective Function}\label{subsec:loss}
\noindent\textbf{Confidence-based Attention. } We segregate the instances into three groups and assign them different attention values. As $\tilde{y} = 0$ always indicates a true negative instance, their attention is assigned $1$. For instances from the positive bag, part of their pseudo labels is clean. We filter out the most noisy ones by forcing their attention to $0$. For the remaining instances, their attention depends on both the prediction and confidence:
\begin{align}
  \hat{a}^{i} =  \left\{\begin{array}{cc}
       c^{i}\times\hat{y}^{i},  & \;\; \text{if} \;\; c^{i}\times\hat{y}^{i}\geq \beta \text{ and } \tilde{y} = 1, \\
        0,  & \;\; \text{if}  \;\; c^{i}\times\hat{y}^{i}< \beta \text{ and } \tilde{y} = 1, \\
        1  & \;\; \text{if}  \;\; \tilde{y} = 0,
      \end{array}
     \right.
\end{align}
where $\beta$ is the threshold that determines how many {\em pseudo} positive samples should be included in the loss function, and we set it as the median of a batch. As indicated by the attention expression, for a pseudo positive instance, we want to focus on those whose prediction that is confidently positive. 

\vspace{2mm}\noindent\textbf{Total loss. }
Given a batch of training samples $\mathcal{D}_{batch}$, we train our framework to minimize the following objective function, which is weighted supervised loss over the instance level prediction and its noisy label $\tilde{y}$. 
\begin{align}
   \min_{\theta} \frac{1}{|\mathcal{D}_{batch}|}\sum_{(\bm{x},\tilde{y})\in\mathcal{D}_{batch}}a((\bm{x}, \tilde{y}), \theta)\times \mathcal{L}_{s}\left(f_{\theta}(\bm{x}), \tilde{y})\right), 
\end{align}
where $a$ is the attention of each instance, $\mathcal{L}_{s}$ is the loss function based on eq\eqref{eq:7}.

\subsection{Uncertainty-Aware Multimodal Fusion}
\label{Uncertainty-aware fusion.}

We incorporate uncertainty awareness in our MIL model that can more accurately detect shapelets (\ie true positive instances), that eventually makes the interpretable ship type prediction (\ie bag level prediction). The MIL model may make less confident predictions if a trajectory is difficult to differentiate. To address this, we propose to fuse the results with other available data modalities. SAR imagery is considered in our work to demonstrate the effectiveness of data fusion. The key assumption is that when multiple data modalities are available, each modality may provide complementary information that may not be available from a single modality alone. 

Feature-level fusion can automatically learn the weights of two modalities but requires the matched training data. However, some data modalities may be limited in their availabilities. We train two independent models from AIS data and SAR images to fully leverage all training instances from two modalities. This is first done separately and is followed by prediction-level fusion. We use a ResNet-18 model for image feature extraction, stacked with a BNN head. The SAR classification network is also uncertainty-aware as the AIS side. 
For a vessel defined by two modalities, $X^\text{AIS}, X^\text{SAR}$,  after acquiring the two models' prediction, the classical way of integrating the predictions is through weighted sum:
\begin{align}
\label{eq:10}
    \hat{Y}_{i} = \lambda\times\hat{Y}^\text{AIS}_{i} + (1-\lambda)\times\hat{Y}^\text{SAR}_{i},
\end{align}
where $\hat{Y}^\text{AIS}$ is the bag level prediction from the AIS model, $\hat{Y}^\text{SAR}$ is the prediction of SAR model, $\lambda$ is a hyper-parameter to balance these two predictions. Ideally, $\lambda$ should be set to favour the model that has overall better performance. However, this requires high-quality validation set to fine-tune the parameter. Another limitation is that the same $\lambda$ is applied to different vessels that lack the flexibility to handle prediction variance over different samples.

Since both the models are Bayesian and uncertainty-aware, we propose to conduct uncertainty-aware fusion to address the limitations outlined above. In particular, $\lambda$ is adjusted automatically to the prediction confidence of each model:
\begin{align}
\label{eq:11}
    \lambda(X^\text{SAR}_{i}, X^\text{AIS}_{i}) = \frac{C(X^\text{AIS}_{i})}{C(X^\text{SAR}_{i}) + C(X^\text{SAR}_{i})},
\end{align}
where $C^\text{AIS}$, $C^\text{SAR}$ denotes the prediction confidence from the AIS model and SAR model, respectively.

%--------------------------------------------------------------

\section{Related Works} 
\label{sec: relatedworks}

% Grabocka \etal~\cite{grabocka2014learning} proposed to optimize a classification objective function to learn near-to-optimal shapelets in sequential data, and achieve true top-K shapelet interactions.

%%%% multivariate sequential data classification
Time series classification problem has been studied for a long time, and the algorithms have developed from traditional shapelets-based~\cite{ye2009time, grabocka2014learning, karlsson2016generalized,wistuba2015ultra,schafer2017multivariate} to recently DNNs based. Ye and Keogh~\cite{ye2009time} first proposed the concept of shapelet. They also developed an efficient algorithm to find the shapelets, which is achieved by employing three novel components: sub-sequence distance early abandon, admissible entropy pruning, and instance reordering. Karlsson \etal~\cite{karlsson2016generalized} introduced the generalized random shapelet forest algorithm, which uses shapelets and tree-based ensembling for univariate and multivariate time series classification. The algorithm uses random instances and shapelets to generate a set of shapelet-based decision trees. Wistuba \etal~\cite{wistuba2015ultra} accelerate the algorithm efficiently by selecting representative patterns from multivariate time series for class discrimination. Sch{\"a}fer \etal~\cite{schafer2017multivariate} extract and filter multivariate features from MTS by encoding context information into each feature, resulting in a small yet very discriminative and useful feature set for MTS classification. 

Recently, some works~\cite{zhang2020tapnet, zhang2020tapnet, karim2019multivariate, dempster2020rocket, fawaz2019deep} have started using DNNs because of their high generalization and capability of dealing with the large volume of data. Tapnet~\cite{zhang2020tapnet} learns the prototype of each class and makes predictions based on the distance between the sample features and the prototype. This approach learns the weights of each training sample, resulting in a representative prototype. Multivariate LSTM-FCNs~\cite{karim2019multivariate} uses an LSTM to encode the temporal relation in time series data and stacks full connected layers for classification. MiniROCKET~\cite{dempster2021minirocket} is an extended version of ROCKET~\cite{dempster2020rocket}, which uses random convolutional kernels to transform a time series into a feature vector. We point more DNNs based methods to the review paper~\cite{fawaz2019deep}.

In this work, we propose to adapt the shapelets idea but identify them from MIL perspective. Moreover, we use powerful DNNs to learn better features from a large volume of data, achieving a superior classification performance.

%-------------------------------------------------------------------------------------------------------
\section{Evaluation and Results}
\label{sec: Evaluation and Results}
We conduct experiments on the real-world AIS and SAR images to validate our proposed framework with respect to the following four research questions: 
\begin{itemize}
    \item {\bf Q1}: Is our TSC model better than the previous methods in terms of ship type classification using the trajectory? This is corresponding to the bag level classification performance. 
    \item {\bf Q2}: Is the model confidence an effective indicator of correct prediction? 
    \item {\bf Q3}: Can the proposed model accurately identify the shapelets?
    \item {\bf Q4}: Is our uncertainty fusion better than the classical prediction fusion method. We do both quantitative analysis and qualitative analysis to answer these questions.  
\end{itemize}

\subsection{Experimental Design}
\label{sec: ExperimentalDesign}

\vspace{-2mm}\noindent{\bf Datasets.} We consider two types of data in our experiments, including AIS and SAR images, where AIS data is used as the primary data modality given its availability. 

\begin{itemize}
    \item \textbf{AIS Data: } AIS is an automated tracking system designed to provide information regarding the ship and its movements to tracking authorities and other vessels. 
    The selected AIS data is provided by the U.S. Coast Guard and is a recompilation of millions of AIS contacts off the coast of North America from 01/01/2020 to 01/31/2020 (approximately 8.1 GB). We use the Maritime Mobile Service Identity (MMSI) to identify the vessel and extract every vessel's trajectory. The data contains ship type values, which allow us to perform a set of ship type classification tasks. 
    After removing some vessels (\ie those with many missing values or trajectories less than $100$ timestamps.), there are a total of $11,054$ vessels that contain $2,220$ cargo, $1,089$ tanker, $542$ fishing, and other type vessels. We use $7,737$ vessels for training and the rest for testing. 
    \item \textbf{SAR Imagery: } We use the OpenSARShip dataset\cite{huang2017opensarship}, which consists of Synthetic Aperture Radar (SAR) images of marine vessels collected from Sentinel-1 Satellite and the corresponding AIS information. There is a total of $5,673$ SAR image instances. This work considers the coarse ship category and trains the model for cargo vs non-cargo classification. We select all the SAR instances with matching MMSI values to make the test set. 
\end{itemize}

\begin{table}[ht!]
\centering
\caption{The bag-level classification performance comparison on three tasks (\%).\label{table1} }
\begin{tabular}{|l|r|r|r|}
\hline
 & \multicolumn{1}{l|}{F-Score} & \multicolumn{1}{l|}{AUC-ROC} & \multicolumn{1}{l|}{AP}  \\ \hline
 & \multicolumn{3}{l|}{Fishing vs Non-fishing}  \\ \hline
Uncertainty-Aware MIL(Ours) & 40.00  & 84.99  & 51.29\\ \hline
Tapnet & 23.49 & 80.57 & 20.28 \\ \hline
LSTM-FCNs &  40.82 & 90.18 & 45.59 \\ \hline
MiniRocket & 13.95 & 68.70 & 63.40   \\ \hline
\hline
 & \multicolumn{3}{l|}{Cargo vs Non-cargo}  \\ \hline
Uncertainty-Aware MIL(Ours) & 50.29 & 87.08 & 60.67\\ \hline
Tapnet & 32.53  & 74.56 & 22.82 \\ \hline
LSTM-FCNs &  34.05 & 80.81 & 64.33 \\ \hline
MiniRocket & 67.73 & 84.00 & 71.10   \\ \hline
\hline
 & \multicolumn{3}{l|}{Tanker vs Non-tanker}  \\ \hline
Uncertainty-Aware MIL(Ours) & 27.26 &  82.52  & 25.08  \\ \hline
Tapnet & 19.05 & 58.33 & 14.08 \\ \hline
% Attention base-MIL &  &  &   \\ \hline
LSTM-FCNs &  13.81 & 75.19 & 24.76\\ \hline
MiniRocket & 54.67 & 87.30 & 63.40   \\ \hline
\end{tabular}
\end{table}

\vspace{-2mm}\noindent{\bf Evaluation Metrics.}
The AIS data is highly imbalanced as the positive instances/bags only occupy a small fraction of the whole dataset. Moreover, in many cases, positive samples are of most interest in the classification task. Consequently, we the F-score, Average Precision (AP) and Area Under the Curve Receiver Operating Characteristic (AUC-ROC) as the evaluation metrics. The results are reported in Table~\ref{table1}.

\vspace{2mm}\noindent{\bf Experiment setup.} The trajectory of each vessel is described by four attributes: 
latitude, longitude, SOG, and COG. We convert the SOG and COG to vertical speed and horizontal speed. We set the length of the sliding window as $100$, which allows DNNs to recognize each instance as it contains sufficient time step, as not exceed the length of a single shiplet. We use the ship type as the bag label. For example, in Fishing vs Non-Fishing, a fishing vessel trajectory is regarded as a positive bag, and all other trajectories are negative bags. Similar treatments are applied to other types of classification tasks. We sample $200$ negative samples and $400$ positive instances to form a batch, avoiding overfit to negative class as positive instances are outnumbered by negative ones. For baseline implementation, We set the maximum length as $100$ because their architectures only allow a fixed temporal length. For Tapnet~\cite{zhang2020tapnet}, we use  $1000$ training trajectories during implementation because it does not allow batch mode training. 

\begin{figure}[t]
\centering
\includegraphics[width=0.8\linewidth]{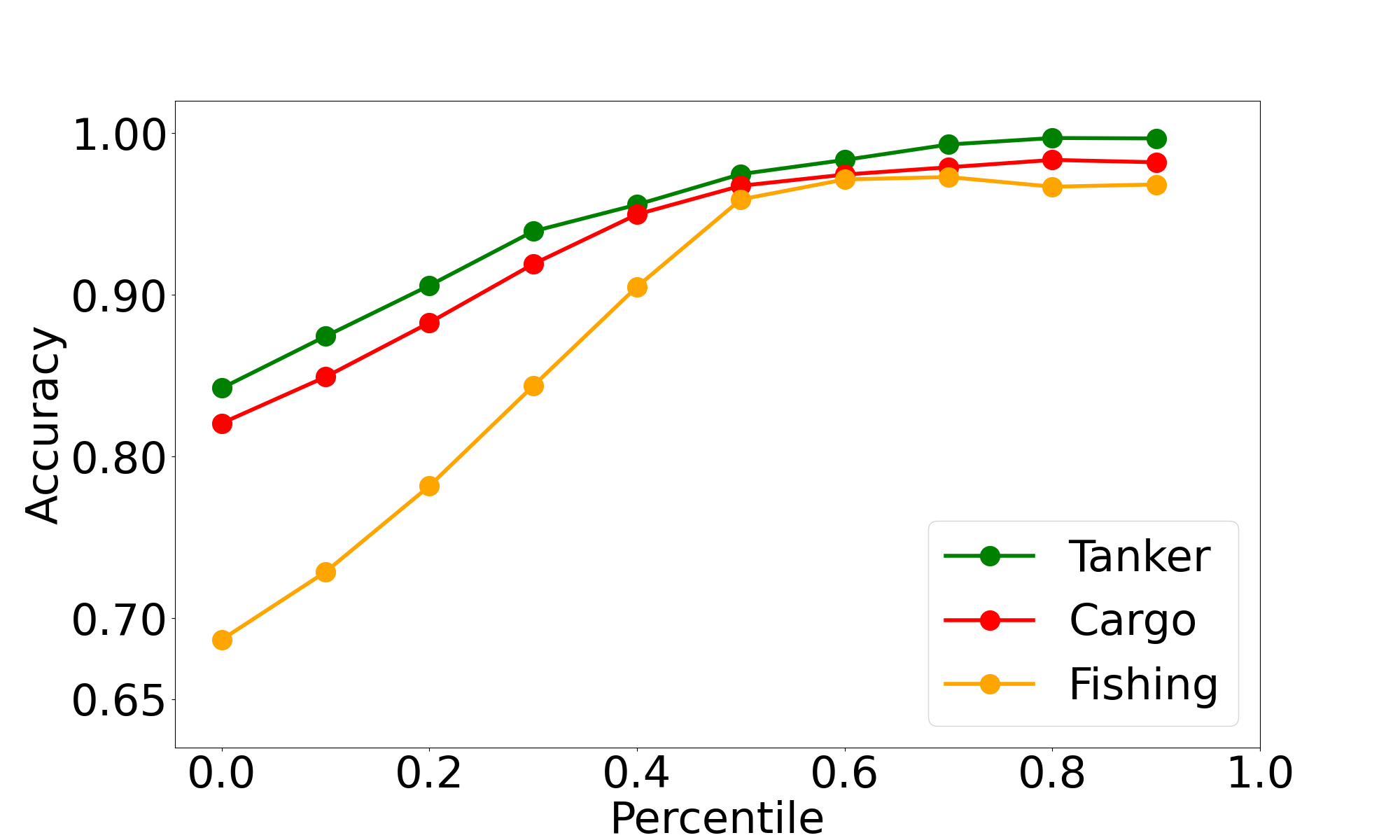}\quad
\caption{Distinguishing between correctly predicted and wrongly predicted labels based on predictive confidence on three vessel type classification tasks. \label{fig:acc_confidence}}
\vspace{-2mm}
\end{figure}
\vspace{-2mm}

\subsection{Experimental Results}
\noindent{\bf Ship type classification.} We evaluate the classification performance with respect to three representative ship types: I) Fishing vs non-fishing vessels,  II) Cargo vs non-cargo vessels, and III) Tanker vs non-tanker vessels. Recognizing these types of ships plays a vital role in the daily affairs of vessel management, monitoring, and scheduling. For example, fishing ship identification can detect IUU fishing activities, and tanker and cargo predictions help secure overseas transportation. We compare our method with Tapnet~\cite{zhang2020tapnet}, Multivariate LSTM-FCNs~\cite{karim2019multivariate}, and MiniROCKET~\cite{dempster2021minirocket}. Details about the baselinees can be found in Section~\ref{sec: relatedworks}.

The result in Table~\ref{table1} shows that our approach achieves consistent and overall the best performance compared to other methods. Among these four methods, TapNet~\cite{zhang2020tapnet} achieves relatively low AUC and low AP score, which  might be due to using a subset of training data. In detecting Tanker or Fishing, Multivariate LSTM-FCNs~\cite{karim2019multivariate} and MiniROCKET~\cite{dempster2021minirocket} archive better performance than our method by a small margin, but their performance is inconsistent and drops a lot in other tasks.  

\begin{figure}[t]
\centering
\includegraphics[width=0.8\linewidth]{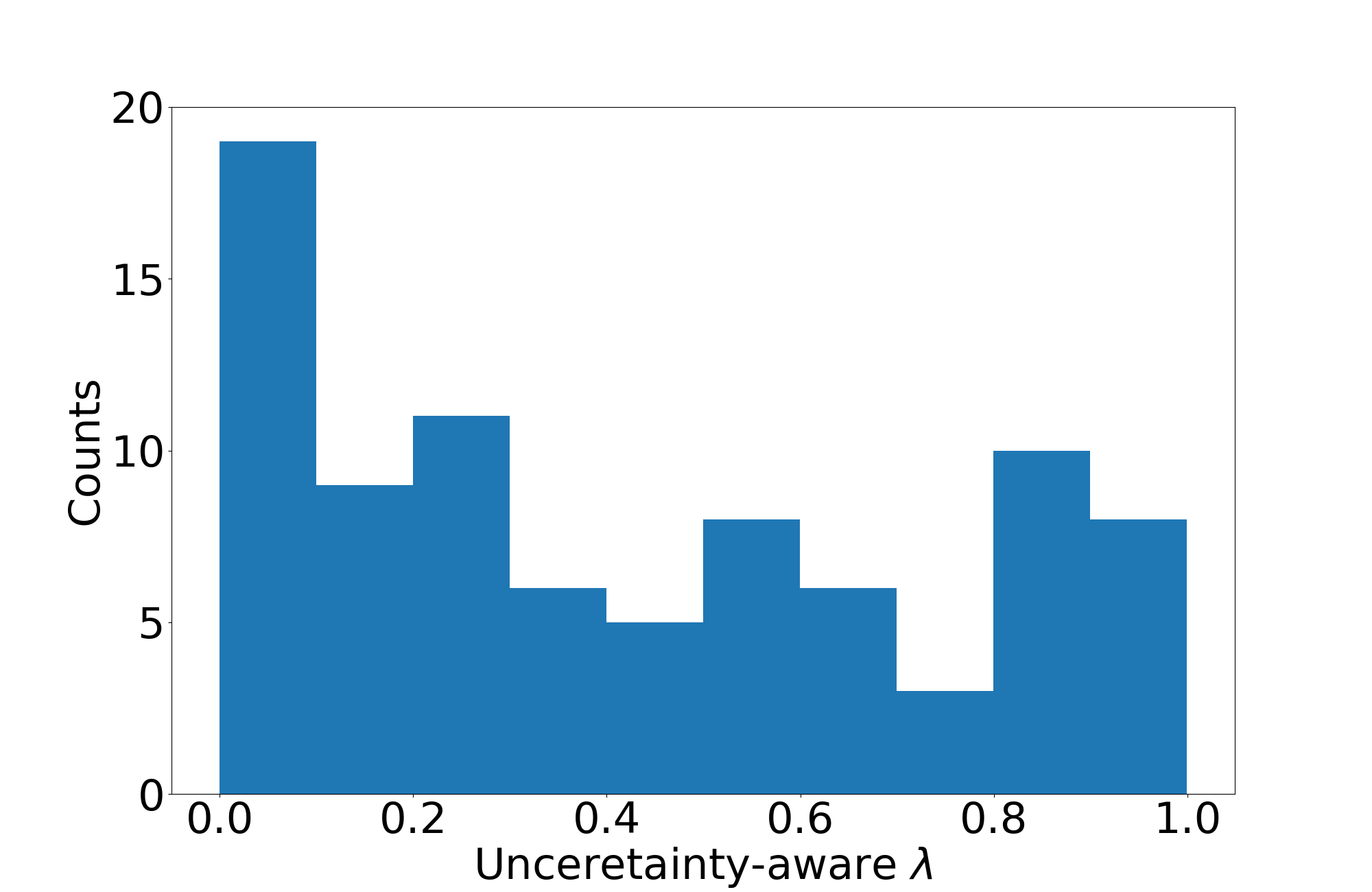}
\caption{Histogram of weight $\lambda$ of all matched samples\label{fig:2.a}}
\vspace{-5mm}
\end{figure}
\begin{figure}[t]
\centering
\includegraphics[width=0.8\linewidth]{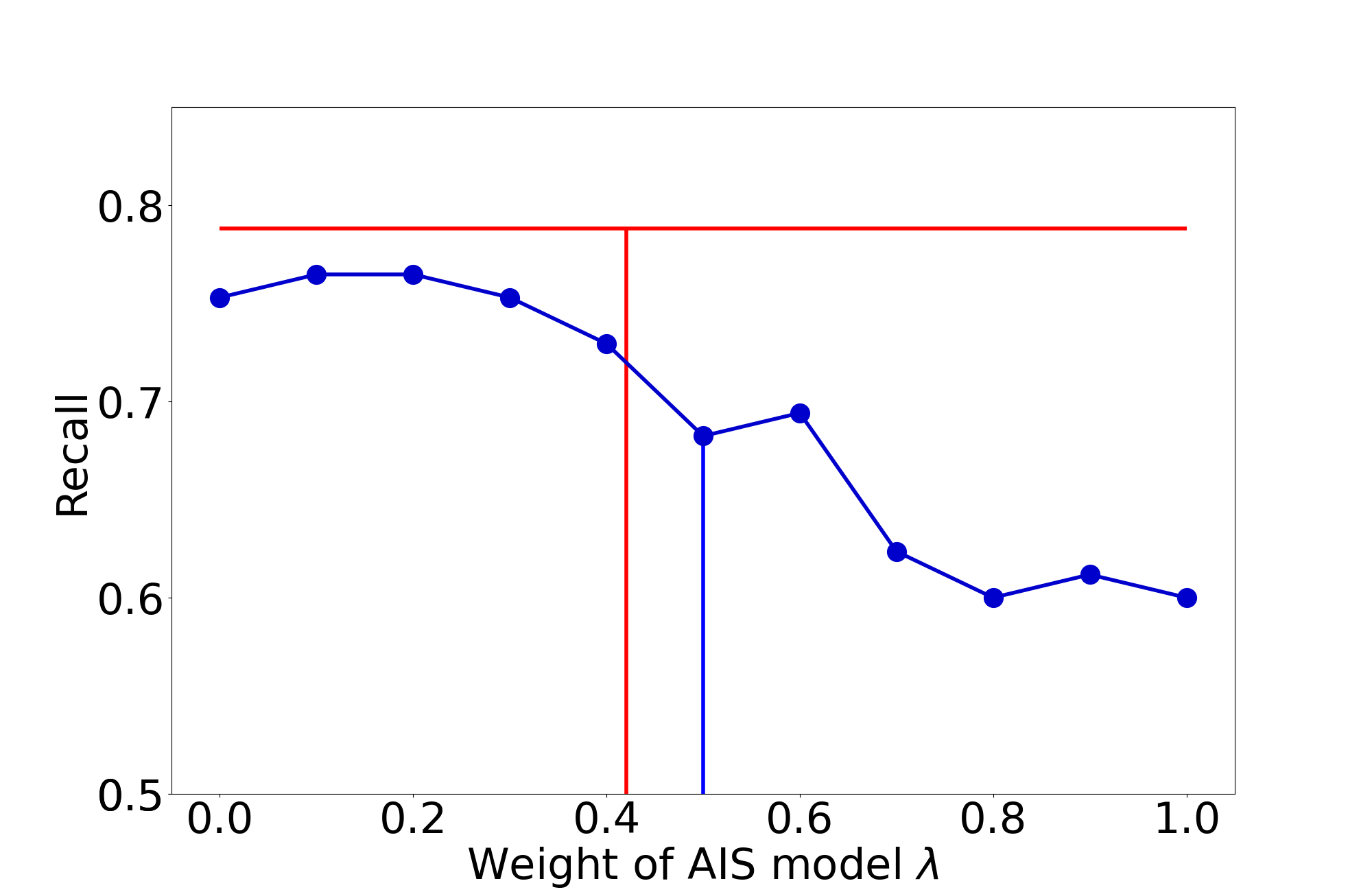}
\caption{Recall of two fusion strategies (blue line and red lines indicate the baseline and uncertainty-aware fusion).\label{fig:2.b}}
\vspace{-4mm}
\end{figure}

\vspace{2mm}\noindent{\bf Uncertainty estimation.} We evaluate the relation between predictive confidence and prediction accuracy. A prediction with low uncertainty (high confidence) is expected to be correct than high uncertainty (low confidence). To assess how strong this trend is, we evaluate the accuracy of prediction at different confidence levels.
We find a set of percentile of the confidence and set these percentages as thresholds, and then we compute the accuracy of predictions whose confidence is above the threshold. Figure~\ref{fig:acc_confidence} shows the relation between threshold and accuracy on three ship type classification tasks. The results consistently indicate that BNN confidence is a reliable indicator of correct predictions.

\vspace{2mm}\noindent{\bf Shapelet identification.} We evaluate the impact of the attention mechanism on detecting shapelets. Since the comparison baselines do not leverage the MIL perspective, they can only perform whole trajectory classification. Thus, we focus on investigating the effect of removing the attention mechanism from the proposed model. We use the Fishing vs Non-fishing classification task to identify the most discriminate instances for fishing vessel trajectory, which are expected to have the navigation status as ``Engaged in Fishing.'' The result shows that the attention mechanism effectively improves the AUC score from 84.11\% to 85.67\%.
%The results in Table~\ref{table:2} show that the attention mechanism effectively contributes to the performance improvement.

\vspace{2mm}\noindent{\bf Data fusion. } We evaluate the impact of different modalities on the model prediction. Figure \ref{fig:2.b} shows that 
%two modalities, SAR and AIS contributes (almost) equally to the model prediction(I assume so, need to see the result to confirm this). However, 
the naive combination of two modalities %(mean/equal weight) 
is ineffective as it results in a marginal performance gain. On the contrary, the proposed uncertainty aware data fusion reaches a good balance between modalities and significantly improves model performance. 

We evaluate the model performance at different degrees of data fusion. Unlike the previous evaluation, the result of data fusion only makes sense for test cases that contain both AIS and SAR information. So we first construct the test dataset by traversing all instances with matching identities (MMSI) from AIS and SAR datasets. Figure \ref{fig:2.b} demonstrates the effectiveness of the uncertainty guided data fusion. The blue curve shows that the tendency of the model recall (y-axis) is affected by fusion coefficient, $\lambda$, where $\lambda=0$ implies the single modality(no fusion) of SAR and the vertical line at $\lambda = 0.5$ implies the equal weight data fusion. Note that the recall curve is sub-optimal since all data instances are forced to adopt the same $\lambda$ for data fusion. The proposed model is more flexible by adapting different $\lambda$s to different data instances. The red horizontal line indicates the recall of the proposed model (0.79) achieved with the averaged $\lambda=0.4$, which is significantly better than equal weight data fusion (0.68). The entire blue curve is under our adaptive data fusion, which outperforms all possible fusion strategies with $\lambda$ fixed to all data instances.
 
In Figure \ref{fig:2.a}, we further show how the proposed approach assigns $\lambda$ values based on data distribution.
%we further show how $\lambda$ values are assigned by the proposed approach distributed on the dataset.
The result shows that the $\lambda$ values are dispersed to different ranges in order to minimize the learning objective. Customizing $\lambda$ for each data instance is the key to data fusion since it allows the model to down weight or switches off the irrelevant or noisy signals from a certain modality according to some prior knowledge. In this work, such prior knowledge is dynamically updated by previously learning experiences and instantiated in uncertainty.
\section{Conclusion}
\label{sec: conclusion}
As the number of data samples, the size of features, and the temporal length in the time series data continue to increase in the big data era, novel challenges arise for large-scale time series classification that makes conventional approaches less effective. In this paper, we tackle the TSC problem from the multiple instance learning perspective by developing a novel uncertainty-aware model to ensure robust model training under noise. In particular, the predicted uncertainty serves as an effective indicator to identify the most probable positive instances, which can lead to improved classification accuracy. When other data modalities are available, uncertainty-aware data fusion can be performed to boost the overall prediction performance further. Extensive experiments conducted on real-world AIS data and the fusion with the SAR images demonstrate the effectiveness of the proposed approach.    

\vspace{-1mm}
\section*{Acknowledgements}
\vspace{-1mm}
This research was supported in part by an ONR award N00014-18-1-2875. The views and conclusions contained in this paper are those of the authors and should not be interpreted as representing any funding agency.

\balance
%\clearpage
\bibliographystyle{abbrv}
\bibliography{main}

\end{document}